\documentclass[IEEEtran,web]{ieeecolor2}

\usepackage[colorlinks]{hyperref}

\usepackage{generic}
\usepackage{cite}
\usepackage{graphicx}
\usepackage{textcomp}
\usepackage{tabularx}
\usepackage{booktabs}
\usepackage{multirow}

\usepackage{microtype} 

\usepackage{tabularx}
\usepackage{color}
\usepackage{amsmath,amssymb,amstext}
\usepackage{multirow}
\usepackage{bbm}
\usepackage{bm}

\usepackage{mathtools}

\renewcommand{\tabcolsep}{2pt}

\usepackage{algorithm}
\usepackage{algpseudocode}

\makeatletter
\algrenewcommand\ALG@beginalgorithmic{\ttfamily}
\makeatother

\DeclareMathOperator*{\R}{\mathbbm{R}}

\DeclarePairedDelimiterX{\infdivx}[2]{(}{)}{#1\;\delimsize\|\;#2}

\usepackage{multirow}

\usepackage{cleveref}

\begin{document}
\title{Explaining deep learning for ECG using time-localized clusters}
\author{Ahc\`ene Boubekki, Konstantinos Patlatzoglou, Joseph Barker, Fu Siong Ng, Ant\^onio H. Ribeiro
\thanks{Ahc\`ene Boubekki is with Physikalisch-Technische Bundesanstalt, Berlin, Germany. Joseph A. R. Barker, Konstantinos Patlatzoglou, and Fu Siong Ng  are with the National Heart and Lung Institute, Imperial College London, United Kingdom. Ant\^onio H. Ribeiro is with the Department of Information Technology, Uppsala University, Uppsala, Sweden.
Corresponding Author: Ahc\`ene Boubekki}
\thanks{}}
\maketitle

\begin{abstract}
Deep learning has significantly advanced electrocardiogram (ECG) analysis, enabling automatic annotation, disease screening, and prognosis beyond traditional clinical capabilities. However, understanding these models remains a challenge, limiting interpretation and gaining knowledge from these developments. In this work, we propose a novel interpretability method for convolutional neural networks applied to ECG analysis. Our approach extracts time-localized clusters from the model’s internal representations, segmenting the ECG according to the learned characteristics while quantifying the uncertainty of these representations. This allows us to visualize how different waveform regions contribute to the model’s predictions and assess the certainty of its decisions. By providing a structured and interpretable view of deep learning models for ECG, our method enhances trust in AI-driven diagnostics and facilitates the discovery of clinically relevant electrophysiological patterns.
\end{abstract}


\section{Introduction}
\label{sec:introduction}

Cardiovascular diseases account for one-third of the deaths worldwide, and the electrocardiogram (ECG) is a major tool in their diagnoses \cite{gbd2016causesofdeathcollaborators_global_2017}. Deep neural networks have recently achieved striking success in the automatic analysis of the ECG. 
Besides providing improved tools for automatically annotating the ECG~\cite{ribeiro_automatic_2020a} that can help assist doctors, machine learning (ML) also allows for the extension of the use of the ECG exam. By leveraging large databases to find new patterns, ML can extract information from the ECG that even trained cardiologists can overlook. Indeed, examples include novel screening algorithms for diseases that traditionally require costly or long exams, such as Chagas disease~\cite{jidling_screening_2023}, left ventricular systolic dysfunction~\cite{sangha_detection_2023}, and electrolyte imbalance~\cite{von_bachmann_evaluating_2024}. 
The automatic analysis of the ECG also brings great promise for prognosis and risk stratification, for instance, of atrial fibrillation~\cite{habineza_end--end_2023}, mortality~\cite{sau_artificial_2024-1}, and cardiometabolic disease~\cite{pastika_artificial_2024}. 

Interpretability plays an important role in allowing AI-driven ECG analysis to reach its full impact. It helps build trust in automatic diagnosis systems and allows doctors to better understand and interact with AI-based support tools. When developing models for new tasks, interpretability can provide insights into electrocardiography itself, helping to bridge the gap between machine learning and medical knowledge. Particularly for tasks that usually cannot be done without machine learning (such as age prediction, risk stratification, and screening for diseases), AI systems can be explored to discover relevant electrophysiological signatures. More generally, being able to inspect and understand how these models work makes it easier to detect potential problems and ensure their reliability in clinical practice.

We propose a novel inspection method for convolutional neural networks (CNNs) for ECG which builds on the intuition behind the widely used and studied hidden markov models~\cite{koski1996modelling,andreao2006ecg}. The equivalent of the \emph{hidden states} are clusters learned in an unsupervised manner based on \emph{observations} that are not the input signal but the internal representations that the encoder creates out of an input. 
The underlying convolutional operations means that the clusters are representing contextualized information processed with respect to the task for which the network was trained. In practice, our method returns a segmentation of the input with clusters specific to a section of the ECG wave and that correlate differently with the labels. Overlying the assignment uncertainty provides extra information on how confident the model is about its representation of each segment of the ECG.

\begin{figure}[t]
    \centering
    \includegraphics[width=\linewidth]{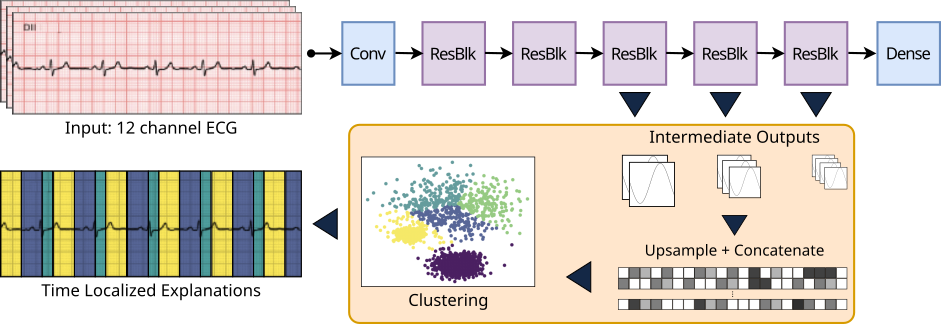}
    \caption{Method illustration. We use time localized cluster for a one-dimensional residual neural network used in ECG automatic analysis.}
    \label{fig:method-visual-summary}
\end{figure}

\section{Time Localized Explanations}
\label{sec:method}

Our method  is inspired by~\cite{boubekki_explaining_2025} but here adapted to the context of multi-channel ECG automatic analysis; we summarize it in Figure~\ref{fig:method-visual-summary}. 
The  cluster using $k$-means the feature activation of one (or several) hidden layer, which we will denote here by ${\bf R} \in \R^{D \times C}$. Here $D$ corresponds to the time dimension and $C$ is the number of channels. Since the order of the time dimension of the internal signals is consistent with that of the input, we can assign timestamps of the input to a specific cluster.
In this section, we first illustrate its use in a common neural network architecture used in the context of ECG and illustrate what kind of visualizations it enables.

\subsection{Concrete example}

To make the example concrete, we make use of the same network architecture as \cite{ribeiro_automatic_2020a}  (we refer to that work for further details), which consists of a residual neural network adapted to uni-dimensional signals (1D-ResNet), with the final output layer adapted for binary classification or regression.

An ECG input ${\bf x}\in \R^{L \times 12}$ of length $L>0$ with $12$ channels is passed through the encoder of the 1D-ResNet consisting of one convolutional layer $l_0$ followed by $5$ residual blocks $l_1,\ldots l_5$.
The internal deep representation produced by the encoder is extracted from the outputs of the last three blocks, ${\bf r}_i \in \R^{d_i \times c_i}$ for $i=3,4,5$.
We chose this combination as trade off between a receptive field consistent with the input time dimension and the representation power of the deep activations. Also, including the output of the last block ensures the incorporation of the information fed to the classifier.
The feature activations are then linearly upsampled to the same length $D=d_3$, scaled, weighted, and finally, concatenated to obtain a matrix ${\bf R} \in \R^{M \times C}$, with $C=c_3+c_4+c_5$. 
The matrix ${\bf R}$, handled as a dataset of $D$ vectors of dimension $C$, is clustered using $k$-means into $K>0$ clusters, yielding the assignment probability matrix ${\bf P} \in \R^{D \times K}$ to the $K$ centroids, computed as the $\operatorname{softmax}$ of the distances to the centroids.
The segmentation ${\bf s} \in [0 \ldots K-1]^D$ derives to the $\operatorname{argmax}$ of the assignment probabilities, while the uncertainty ${\bf u} \in \R^D$ corresponds to their entropy. 
The final clustering, or segmentation, can, therefore, be interpreted as a quantized explanation of the encoder's internal representation of the input. 
A pseudo-code of these operations is given in Algorithm~\ref{algo:method} and a schematic representation in Figure~\ref{fig:method-visual-summary}.

\begin{algorithm}[!t]
    \caption{Pseudocode for time-localized cluster in the last three layers of the network.}
    \label{algo:method}
    \small
    \begin{algorithmic}[1]
        \Require ECG ${\bf x}$ of length L, nb. of clusters K
        \State z = l${}_2$ $\circ$ l${}_1$ $\circ$ l${}_0$(${\bf x}$)
        \State {\bf R} = [] 
        \For {i = 3 to 5}
            \State z = l${}_i$(z)
            \State r = linear\_upsample(z, D)
            \State r = r / norm(r, ord=2, axis=1)
            \State r = r / (1 + d${}_{i}$) 
            \State {\bf R}.append(r)
        \EndFor
        \State {\bf R} = concatenate({\bf R}, axis=1)
        \State {\bf P} = KMeans(K).fit\_predict\_proba({\bf R}) 
        \State {\bf s} = argmax({\bf P}, axis=1)  
        \State {\bf u} = entropy({\bf P}, axis=1) 
        \State \Return {\bf s}, {\bf u}
    \end{algorithmic}
\end{algorithm}

\begin{figure}
    \centering
    \includegraphics[width=\linewidth]{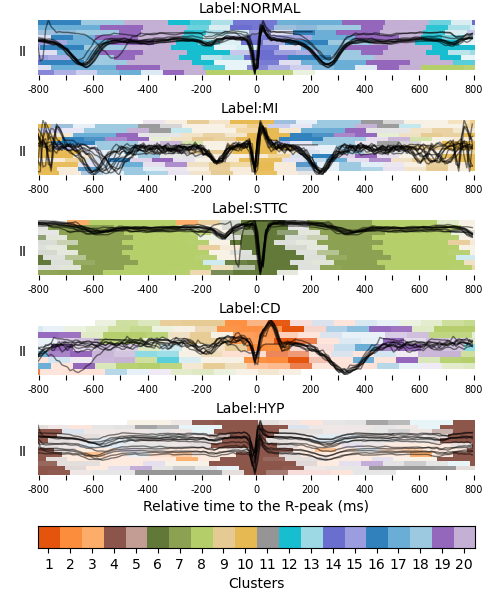}
    \caption{Time localized explanations ($K=20$) with uncertainty (opacity) of ECGs with different conditions.}
    \label{fig:example}
\end{figure}

\subsection{Visualization}
In Figure~\ref{fig:example}, we show explanations with uncertainty (opacity) for each label of PTB-XL dataset~\cite{wagner_ptbxl_2020}. 
The curves in the foreground are the beats in the same ECG centered on their QRS complex detected using NeurKit~\cite{makowski_neurokit2_2021}. 
The x-axis indicates the relative time in milliseconds before and after the R-peaks.
The background is made up of stacked explanations of the several sinuses of the ECG. 
This means, for example, that the ECG of the top figure contains eleven QRS complexes. 
The colors correspond to the clusters (here $K=20$), and the opacity indicates the uncertainty of the explanation (the lower the opacity, the more uncertain the assignment).

We make three hypotheses based on the Figure~\ref{fig:example}.
First, different heart conditions seem to leverage different clusters. 
Second, the clusters seem specific to different phases of the sinus rhythm. 
Finally, the explanation tends to be more certain around the QRS complexes. 

These hypotheses are supported by the visual patterns in Figure 2 when interpreted through a clinical lens. The model assigns distinct clusters to different waveform components depending on the diagnostic label, suggesting its internal representations are class specific rather than generic. Cluster boundaries align with key electrophysiological segments such as the P-wave, QRS complex, and T-wave, indicating a structured, time-localized decomposition of the ECG related to physiology that it has not been explicitly taught. In normal ECGs, clusters are most distinct, consistent, and confident, likely reflecting underlying physiological stability and signal regularity as well as an over representation of normal ECGS on the training set. In myocardial infarction (MI), however, confident and concentrated clustering is mostly around the QRS and early ST segment, reflecting ischemic changes in depolarization and repolarization. ST/T changes (STTC) display prominent clustering around the T-wave, especially its peak and offset, consistent with nonspecific repolarization abnormalities. Hypertrophy (HYP) examples exhibit dense clustering across the QRS and early T-wave, matching expected voltage criteria and repolarization strain in left ventricular hypertrophy. In contrast, conduction disturbances (CD) show fragmented and uncertain clustering around the QRS complex. This likely reflects the heterogeneity of the CD class, which includes both left and right bundle branch blocks, conditions that differ markedly in morphology and pathophysiology. The model's uncertainty clustering mirrors this intra-class variability. Similarly, heterogeneity between the P-wave and QRS in CD cases likely arises from the inclusion of diverse atrioventricular (AV) block phenotypes, which vary in atrioventricular conduction timing and degree of dissociation. Here, representation entropy highlights not only model uncertainty but also label inconsistency or granularity issues. This enhances the value of the method for both interpretation and quality control, identifying where labels may obscure physiologically distinct subtypes, which is critical for safe and effective clinical deployment.

\section{Experimental Setting}
\label{sec:setting}

\subsection{Datasets}
\noindent
We use two datasets to develop and evaluate our models.

\textbf{PTB-XL}~\cite{wagner_ptbxl_2020} is a dataset consisting of 21,799 12-lead ECG records collected between 1989 and 1996 from 18,869 different patients using devices from Schiller AG. The dataset is split into ten folds. The backbone ResNet is trained on the nine first and the explanation method on the last one (2,198 instances). 
PTB-XL covers a larger range of diagnostic statements than the CODE dataset, with additional types of arrhythmias and diagnostics annotated.

\textbf{CODE-15\%}~\cite{ribeiro_automatic_2020a} is an open available subset of the CODE dataset\footnote{Link: \url{ https://doi.org/10.5281/zenodo.4916206}}, containing 345,779 exams from 233,770 patients.
The Clinical Outcomes in Digital Electrocardiography (CODE) dataset was developed with the database of digital ECG exams of the telehealth network of Minas Gerais, Brazil. and a detailed description of the cohort can be obtained at~\cite{ribeiro_automatic_2020a}. 
Here, to evaluate the explanations capabilities, we use a holdout evaluation set of 5,779 samples from CODE-15\% .

\subsection{Tasks}

We focus our analysis on the explanations for two tasks: (i) classification of ECG abnormalities and (ii) age prediction. The classification of ECG abnormalities in PTB-XL is detailed in~\cite{strodthoff_deep_2020}, while the age prediction task in the CODE dataset is described in~\cite{lima_deep_2021}.

These tasks serve distinct purposes. Classifying ECG abnormalities is a well-established problem with known ECG characteristics, making it a suitable benchmark for evaluating our interpretation algorithm. In contrast, age prediction extends beyond traditional electrocardiography, providing an opportunity for our interpretation techniques to uncover underlying patterns and mechanisms.

The model trained set for multilabel classification using the diagnostic superclass: 
Normal (NORM), Myocardial Infarction (MI), ST/T Change (STTC), Conduction Disturbance (CD) and Hypertrophy (HYP).
The model returns on the test an average label-wise accuracy and AUROC of $86.4\% \pm 2.1$ and $79.2\% \pm 4.4$, respectively. For age-prediction on CODE-15\%, we use the openly available model's weights which achieves a mean absolute of $8.72 \pm 7.25$ years.

\section{Explanations of Abnormalities}
Below, we describe experiments conducted using our methods to gain insights from the models outlined above using $K=20$ clusters. We begin by focusing on the multilabel classification problem, and in~\Cref{explaining-age}, we present analogous experiments for the age prediction model. For the multilabel classification problem, we provide in the appendix equivalents of Figures~\ref{fig:correlation} and \ref{fig:wave} for $K=5,\:10,\:15$ and $20$.

\subsection{Explanation and Classes}

\begin{figure}[!h]
    \centering
    \includegraphics[width=\linewidth]{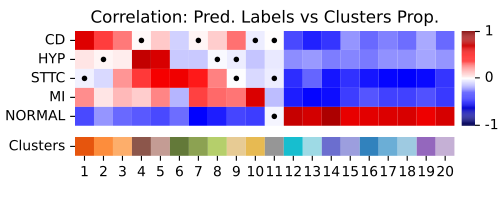}
    \caption{Pearson correlation between the predicted labels and clusters' proportions in an explanation. Dots indicate p-value larger than $0.05$ (non-correlation).}
    \label{fig:correlation}
\end{figure}

In Figure~\ref{fig:correlation}, we depict the Pearson correlation matrix between the predicted labels and the proportion of each cluster in the sequences.

Most of the correlations are statistically significant, which partly satisfies our first hypothesis that the clusters are specific to the labels.
Indeed, clusters 12 to 20 are dedicated to ``normal" ECGs, while the others are distributed between the different labels. Note that while cluster 11 only slightly correlates with one label, other clusters (e.g., cluster 4) correlate strongly with several conditions.

\subsection{Explanation and ECG Key Points}

\begin{figure}[!h]
    \centering
    \includegraphics[width=\linewidth]{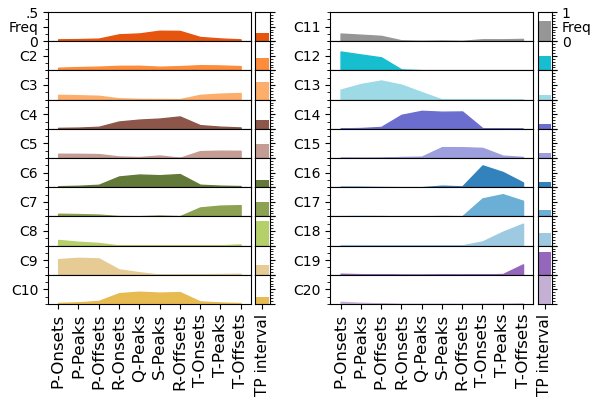}
    \caption{Frequency of ECG key points within each cluster. \emph{Note: To improve visualization, the TP interval is shown on a 0 to 1 scale, while all other values are shown on a 0 to 0.5 scale.}}
    \label{fig:wave}
\end{figure}

In Figure~\ref{fig:wave}, we depict the cluster frequency of the 8-millisecond segments containing the ECG key points.

The plots confirm the specialization of each cluster for the different parts of the wave. 
Since more clusters are specific to normal ECG, the waves are decomposed into more clusters. 
This specialization is particularly noticeable for the T-wave, which is the captured by clusters 16, 17, and 18, focusing respectively on the onset, peak, and offset of the wave. 
In the case of the clusters correlating with some medical conditions, the partition of the signal is simpler: Some clusters focus on the QRS complex, and others on the rest of the signal. 
Cluster 11, which correlates the least with the labels (Figure~\ref{fig:correlation}), appears mostly outside ($68\%$ of the segments) the sinuses (TP interval). Clusters 8, 19 and 20 are even more frequent within the TP interval ($84\%$, $78\%$ and $94\%$, respectively), yet they correlate with some labels.

Overall, the explanations reflect a clinical logic: where the ECG is stable and well-formed, the segments are finer; where pathology introduces noise or variability, segmentation becomes coarser but potentially more label-driven. This phase-aware, label-specific clustering provides a mechanistic view of how the model processes ECG signals, aligning with the clinician’s focus on waveform structure while remaining unconstrained by the century-old, macroscopic PQRST framework that often oversimplifies the complexity of electrocardiographic physiology. Instead of relying solely on discrete fiducial points, the model captures continuous, data-driven representations that may reflect latent physiological patterns beyond current clinical resolution.

\subsection{Explanation and Prediction}\label{sec:pred}

The specialization of the clusters suggests that they could be used for prediction.
To demonstrate this idea, we train a random forest classifier on the raw signal (RF signal) and on the proportion of each cluster in each ECG (RF cluster). For the latter, the explanations are trained on the same $500$ ECGS of the test set. 
We report in Table~\ref{tab:prediction} the average scores computed over three repetitions of 5-fold cross-validations on the whole test set.
For comparison, we include the performance of the 1D-ResNet with respect to the true labels.

\begin{table}[!h]
    \centering
    \caption{Performance of a random forest classifier using raw data (RF signal) or the clusters' proportions (RF cluster) with respect to the true labels (Labels) or to the 1D-ResNet's predictions (Pred.).}
    \begin{tabular*}{\linewidth}{@{\extracolsep{\fill}}lccccc}
        \toprule
                    & 1D-ResNet & \multicolumn{2}{c}{RF signal}    & \multicolumn{2}{c}{RF cluster} \\ 
                    & Labels    & Labels    & Pred.             & Labels    & Pred. \\ \midrule
         Accuracy   & 0.86      & 0.77      & 0.79              & 0.87      & 0.93 \\
         Precision  & 0.71      & 0.48      & 0.51              & 0.77      & 0.87 \\
         Recall     & 0.68      & 0.17      & 0.2               & 0.6       & 0.75 \\
         AUROC      & 0.79      & 0.56      & 0.57              & 0.77      & 0.86 \\
         F1         & 0.69      & 0.22      & 0.25              & 0.66      & 0.8 \\
         \bottomrule
    \end{tabular*}
    \label{tab:prediction}
\end{table}

Random Forest trained on the clusters' proportions ($\operatorname{RF cluster}$) is better at predicting the true labels than when trained on the raw data ($\operatorname{RF signal}$). 
Interestingly, that same $\operatorname{RF cluster}$ performs on par with the 1D-ResNet on the true labels with respect to all the metrics. This observation is confirmed by the similarity of their predictions (last column). 
This suggests that the classifier of the 1D-ResNet leverages information akin to distribution of the segments of the encoder's explanations.

\subsection{Explanation Uncertainty}

The uncertainty of the encoder's explanation does not necessarily match the importance of the feature with respect to the predictions. The difference is that the latter includes the classifier and computes the importance of the feature with respect to the prediction. In our method, the predictions are not involved as the uncertainty corresponds to that of the cluster assignments of activations extracted from the encoder only. We study the dissimilarities by comparing the uncertainty of our method with the saliency maps of Grad-CAM~\cite{selvaraju_gradcam_2020}, which has been used for ECG~\cite{gustafsson_development_2022, lima_deep_2021}.

\begin{figure}[!h]
    \centering
    \includegraphics[width=\linewidth]{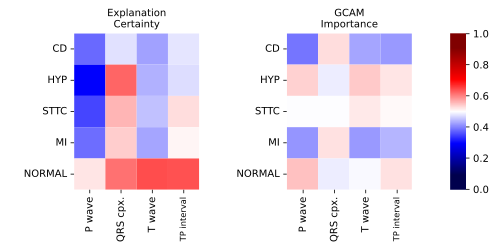}
    \caption{Explanations certainty compared to Grad-CAM importance with respect to the different phases of the ECG.}
    \label{fig:uncert}
\end{figure}

Grad-CAM saliency maps are computed for each label at the output of the first convolution layer before merging the channels with a $\max$. To enhance the contrast, the $\log$ of the resulting map is standardized before being applying the $\operatorname{sigmoid}$ function with a temperature of $10$. The same transformation is applied to the uncertainty maps of the encoder's explanations.
Figure~\ref{fig:uncert} depicts the average uncertainty/importance for both methods per predicted label. Here, the wave key points are merged per complexes. 
A visual comparison is depicted in Figure~\ref{fig-apx:uncert} in the Appendix.

The striking differences between the two heatmaps confirm that uncertainty of the representation explanations and feature importance are two different things. 
Except for CD, the representation of the around the QRS complex are quite certain than the rest, which is in line with our hypothesis. An interpretation is that the encoder is quite certain at recognizing the QRS complexes.
Interestingly, the explanations of the ``normal" ECGs are quite certain (red) all over the signal, while for the other labels, the uncertainty is high (blue) outside the QRS complex. This over confidence of the encoder for that specific label might be due to the over representation of "normal" ECGs in the dataset.
As for Grad-CAM, except for CD and MI, parts of the waves outside the QRS complex are deemed more important for the predictions than the QRS. 
Note that the relationship between the uncertainty of the explanations and the feature importance is not clear. Further investigation is needed on that point.

\subsection{Explanation of Misclassifications}

\begin{figure}[!h]
    \centering \small
    a. Expert label error with concordant model predictions \\
    \includegraphics[width=\linewidth]{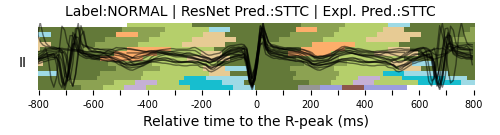}\\ \vspace{.5em}
    b. Expert label error with discordant model predictions \\
    \includegraphics[width=\linewidth]{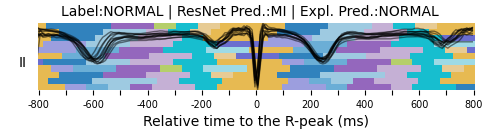}\\ \vspace{.5em}
    c. Limitations of global labels for transient ECG morphologies \\
    \includegraphics[width=\linewidth]{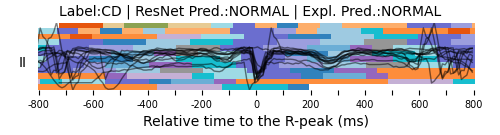}\\ \vspace{.5em}
    \caption{Stacked explanation plots of misclassified examples with the 12-lead shown in Figures \ref{fig:misclass_full_1} and \ref{fig:misclass_full_2} in the Appendix.}
    \label{fig:misclass}
\end{figure}

We use the misclassified examples in Figure~\ref{fig:misclass} to illustrate how encoder-based explanations can provide clinical insights as well as reveal limitations of the model and methodology in cases where predictions diverge from ground truth labels. For each example, we indicate the predictions from both the 1D-ResNet and the $\operatorname{RF cluster}$. Unstacked 12-Lead ECGs with class probabilities and explanations most relevant to clinical readers are plotted in Figures~\ref{fig:misclass_full_1} and~\ref{fig:misclass_full_2} in the Appendix.

In the first example the ECG is labeled as normal but predicted as ST and T-wave changes (STTC) by both the ResNet and the RFCluster. Upon clinical review of the 12-lead in Figure~\ref{fig:misclass_full_1}.a, the tracing demonstrates pathological Q waves in the chest leads with reciprocal early repolarisation, T-wave flatting and inversion laterally. This means the PTB-XL labeler has been subject to human error,  mischaracterising the ECG of someone with a previous myocardial infarction (MI) and residual STTC as normal. This highlights the model utility as an additional layer of safety in ECG interpretation to nudge clinicians during interpretation.

The second example is a label of normal but this time with discordant model predictions; the ResNet predicts MI and RFCluster predicts normal, but with interesting granularity of the clusters demonstrated in Figure~\ref{fig:misclass}.b. Upon clinical review this is also an example of pathological Q waves, early repolarisation and more extreme T wave inversion possibly suggestive of active ischaemia requiring symptomatic correlation, and not available within the dataset. Here the ResNet prediction of MI, and confidently at $99.9\%$, which is not unreasonable. Upon review of the clusters in Figure\ref{fig:misclass_full}.b we can see the yellow clusters associated with MI are mostly consistent around the pathological QRS but with the normal clusters located elsewhere. This again shows the utility of our method to highlight portions of the ECG for clinical review. 

Finally, Figure~\ref{fig:misclass}.c illustrates a case where both the ResNet and RFCluster predict a normal label, which does not align with the ground truth label of Conduction Disease (CD). Upon review of the full 12-lead ECG provided in the supplement (Figure~\ref{fig:misclass_full_2}), ventricular ectopic beats are clearly present and are appropriately highlighted in orange—corresponding to the cluster associated with CD. However, since only three ectopic beats occur across the 10-second recording, the model is not necessarily incorrect in labeling the overall ECG as normal, as the majority of the tracing is indeed normal. This example is neither a clinical oversight nor a mislabel, but rather a limitation inherent to assigning a single label to an entire 10-second strip in the presence of transient ECG morphologies. It further underscores the utility of the explanations in guiding clinician attention to relevant features during ECG review.

\subsection{Ablation Study}

Our method relies on two hyperparameters: the number of clusters and the amount of data on which the clustering is learned. 
We leverage the results of Section~\ref{sec:pred} and use the performance of a random forest classifier trained on the explanation to evaluate the influence of these hyperparameters.

\subsubsection{Training size}
We train the explanations on random subsets of varying given size, and $K=20$.
The scores are computed as the average of three repetitions of 5-fold cross-validations on fold 10 of PTB-XL. The consistency of the metrics over the different settings as reported in Table~\ref{tab:abl-ndata} demonstrates the robustness of the explanations with respect to the amount of data our method is trained on.
\begin{table}[!h]
    \centering
    \caption{Minimal impact of the training data size on the prediction capabilities of the explanations.}
    \begin{tabular*}{\linewidth}{@{\extracolsep{\fill}}lccccccc}
        \toprule
Train. size & 50 & 100 & 200 & 500 & 1000 & 1500 & 2198\\ \midrule
Accuracy & 0.87 & 0.87 & 0.87 & 0.87 & 0.87 & 0.87 & 0.87\\
Precision & 0.77 & 0.76 & 0.76 & 0.77 & 0.77 & 0.77 & 0.76\\
Recall & 0.6 & 0.59 & 0.6 & 0.6 & 0.62 & 0.61 & 0.61\\
AUROC & 0.76 & 0.76 & 0.76 & 0.77 & 0.77 & 0.77 & 0.77\\
F1 & 0.66 & 0.65 & 0.66 & 0.66 & 0.68 & 0.67 & 0.67\\
         \bottomrule
    \end{tabular*}
    \label{tab:abl-ndata}
\end{table}

\subsubsection{Number of Clusters}
We train now our method on the same random subset of size $500$ and let the number of clusters $K$ vary.
The scores are computed as the average of three repetitions of 5-fold cross-validations and reported in Table~\ref{tab:abl-nclust} 
If the accuracy seems to be invariant to the number of clusters, other metrics are more sensitive. For example, Precision, i.e., the proportion of labels correctly predicted, increases with the number of clusters.
Overall, the performance is the worst for $K=5$ and stabilizes after $K=20$. We provide more details in \Cref{fig-apx:ablk,fig-apx:ablk2} in the Appendix.
\begin{table}[!h]
    \centering
    \caption{A finer clustering yields .}
    \begin{tabular*}{\linewidth}{@{\extracolsep{\fill}}lccccc}
        \toprule
Nb. Clusters ($K$) & 5 & 10 & 20 & 50 & 100\\ \midrule
Accuracy & 0.82 & 0.85 & 0.86 & 0.86 & 0.86\\
Precision & 0.59 & 0.71 & 0.77 & 0.76 & 0.78\\
Recall & 0.48 & 0.56 & 0.58 & 0.57 & 0.56\\
AUROC & 0.69 & 0.74 & 0.76 & 0.75 & 0.75\\
F1 & 0.52 & 0.62 & 0.65 & 0.63 & 0.62\\
         \bottomrule
    \end{tabular*}
    \label{tab:abl-nclust}
\end{table}

\subsection{Explaining Age Prediction}
\label{explaining-age}

We unroll here a similar explanation protocol but applied to explain the prediction of the age of a patient based on an ECG~\cite{lima_deep_2021}. The base model is the same 1D-ResNet but with a single output classifier trained to regress the true age. The explanations are again computed using the outputs of the last three blocks and $K=20$ clusters.

\subsubsection{Correlations} 
\begin{figure}[!h]
    \centering \small
    \includegraphics[width=\linewidth]{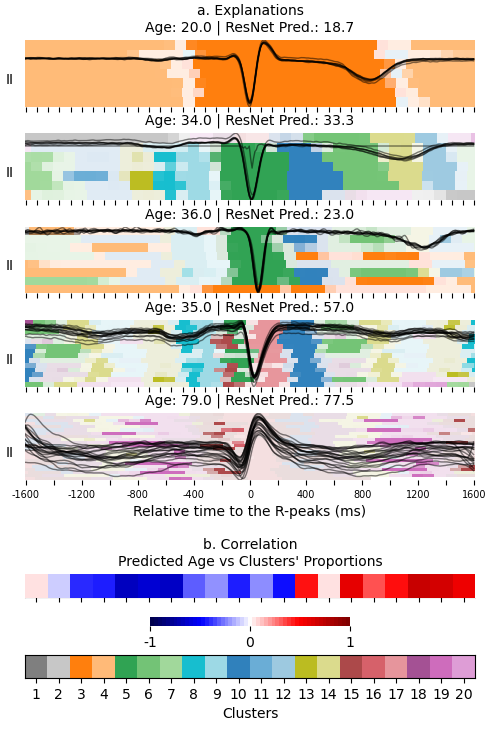}\\
    \caption{Time localized explanations ($K=20$) of ECGs for the task of age regression.}    \label{fig:age_example}
\end{figure}

In Figure~\ref{fig:age_example}, we show explanations along with the Pearson's correlation between the clusters' proportions and the predicted age (all p-value are smaller than $0.05$). We selected five examples to highlight different behaviors. The first three examples cover the age range of the population for which the model correctly predicts the age.
The young person's ECG is mostly decomposed into two clusters (3 and 4), while the others involve more clusters. This may be due to the imbalance of the dataset toward very young patients.

The last two examples are patients in their thirties for whom the model predicts a younger (fourth example) or older (fifth example) age. Note the presence of the orange clusters 3 and 4 in the former's ECG explanation (age 36 predicted 23), while the latter's (age 35 predicted 57) has the end of its QRS complex assigned to cluster 17 (pink), which correlates with older age range.

\subsubsection{Age Distribution} 
\begin{figure}[!h]
    \centering \small
    \includegraphics[width=\linewidth]{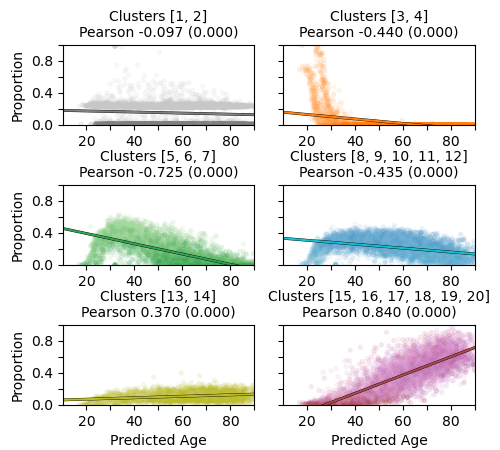}\\
    \caption{Distribution of the proportion of groups of clusters over the predicted age. In addition, we report the Pearson's correlation scores and regression lines.}
    \label{fig:age_prop}
\end{figure}

We continue the analysis with the evolution of the proportion of groups of clusters with respect to the predicted age shown in Figure~\ref{fig:age_prop}. Clusters with similar plots are grouped together, which matches the color scheme of Fig.~\ref{fig:age_example} and \ref{fig:age_wave}. 

Clusters 1 and 2 (gray) corresponds to the padding at the extremities of the timeseries and are thus either present or not. As expected, the orange clusters (3 and 4) are very specific to very young age and become almost absent after the age of 40. 
The distributions of the green and blue clusters over the predicted age form bridges. The proportion of the green group increases quickly to a proportion of up to $0.6$ reached around the age of 35 and then decreases. It can be absent in some patients who are predicted to be older than 50 years old. 
The blue group presents a more moderate increase and decrease and remains present in most patients after their twenties until the late sixties. 
The last two groups correlate positively with the predicted age. While the proportion of the yellow cluster increases slowly with predicted age, the red ones cover more than $30\%$ of the ECG after the age of $60$.

The specificity of the clusters to an age range is remarkable. Here again, the number of clusters associated to a certain age range matches the frequency of patients within that range.

\subsubsection{ECG Key Points} 
\begin{figure}[!h]
    \centering \small
    \includegraphics[width=\linewidth]{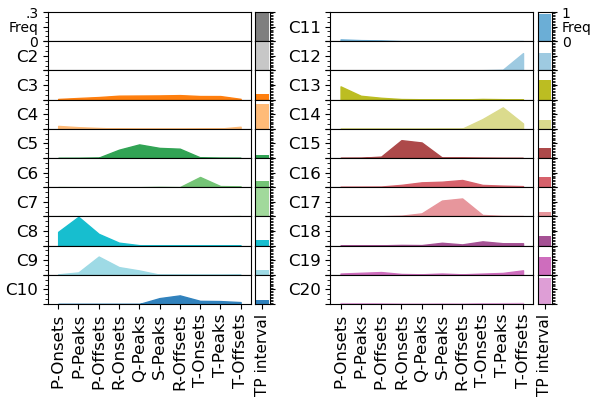}\\
    \caption{Frequency of ECG key points within each cluster. \emph{Note}: To improve visualization, the TP interval is shown on a 0 to 1 scale, while all other values are shown on a 0 to 0.3 scale.}
    \label{fig:age_wave}
\end{figure}

Similar to the previous task, the clusters are specific to a certain location of the signal, as shown in Figure~\ref{fig:age_wave}. Clusters 1 and 2 are specific to the padding at the beginning and the end of the ECG, hence they never overlap a key point. 
Here again, if more clusters specific to the same age range, they focus on different parts of the ECG.
Clusters 3 and 4, which are predominant among patients predicted as young, appear to be mutually exclusive and complementary. The former captures the sinuses, while the latter focuses on the rest of the signal. 

\subsubsection{Uncertainty} 
\begin{figure}[!h]
    \centering \small
    \includegraphics[width=.7\linewidth]{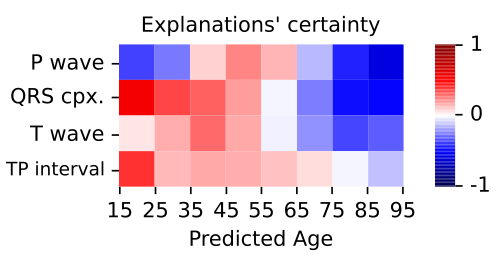}\\
    \caption{Explanations certainty with respect to the different phases of the ECG stratified along the predicted age range.}
    \label{fig:age_uncert}
\end{figure}

Finally, we complete the study of localized explanations of the age regression model with an analysis of the uncertainty of the explanation. In Figure~\ref{fig:age_uncert}, we represent the uncertainty of the explanation stratified along 5-year age groups. Red and blue shades indicate more or less certain explanations. Here, we do not include Grad-CAM importance, since it returns inconsistent results for this regression problem.

Overall, the explanations become less certain as the age increases. The representation of the QRS complexes is the part that sees the certainty of its representation decrease the fastest with age.

Interestingly, in the range 15-25 years old, the explanations of the QRS complex and outside the sinuses are very certain. This age range relies mostly on just two clusters covering almost exclusively these two areas. This result confirms that the encoder is quite certain in distinguishing the QRS complexes from the rest.

\section{Discussion}

In this work, we presented a novel interpretability method for deep learning applied to ECG analysis. By extracting time-localized clusters from internal network representations, we provide intuitive visualizations and quantifiable measures of uncertainty. This approach enhances the transparency of ECG neural networks, allowing for a deeper understanding of how different waveform segments contribute to classification and regression tasks. This is a post hoc method that allows analysis of most common neural network architectures, even when interpretability was not considered during training. We believe it is a powerful tools for examining model behavior in the context of biomedical signal processing. 

Results from both the classification task (PTB-XL) and the regression task (CODE15\% for age prediction) demonstrate the method’s effectiveness in uncovering meaningful patterns in ECG signals. In the classification of common diagnoses, we observed strong correlations between cluster assignments and diagnostic labels, Some clusters are predominantly associated with specific conditions (e.g., STTC or Myocardial Infarction), while others are more generic, as illustrated in \Cref{fig:correlation}.  This is further supported by the high performance of predictive models that use the clusters as input features . Notably, the cluster representation yields strong performance even when learned on small subsets of the data, as shown in \Cref{tab:abl-ndata}.

Furthermore, our findings show that clusters specialize not only in clinical labels but also in waveform (\Cref{fig:wave}). Interestingly, even when the model fails to predict the correct label, the explanation can highlight possible causes, such as ambiguous waveforms or conflicting signal patterns, which might even suggest labeling inconsistencies~\Cref{fig:misclass}. This highlights the value of our method for both model diagnostics and dataset quality control. This has been an important use of explainability methods in medical imaging, such as detecting how state-of-the-art melanoma models relying on spurious features~\cite{winkler_association_2019}.

By visualizing the entropy of cluster assignments (\Cref{fig:uncert,fig:age_uncert}), we introduce a novel approach to assess model confidence in the information actually used by the classifier—unlike post-hoc saliency methods such as Grad-CAM. Our results suggest that the encoder’s uncertainty profile captures richer and more nuanced information, revealing distinct patterns of uncertainty across different abnormalities (\Cref{fig:uncert}), whereas Grad-CAM tends to yield more uniform attention maps. Notably, representation uncertainty increases with patient age (\Cref{fig:age_uncert}), raising important questions about model robustness in clinically challenging scenarios.

The predictive capacity of the explanations further strengthens the case for their informativeness. We demonstrated that a simple random forest trained on cluster proportions outperforms one trained on raw signal data, and even approximates the performance of the full 1D-ResNet (\Cref{tab:prediction}). This suggests that the encoder’s learned features, as captured by the time-localized clusters, encode most of the discriminative information used for classification. It also points to the potential of using our explanations for lightweight, interpretable models suitable for resource-constrained settings (\Cref{tab:abl-ndata}).

Ablation studies in \Cref {tab:abl-nclust} confirmed the robustness of our method with respect to the number of clusters. While performance plateaus beyond 20 clusters, finer granularity does not degrade performance and might support more detailed downstream analyses. These findings highlight the practicality of the method and its low sensitivity to hyperparameter selection, making it well-suited for real-world deployment where parameter tuning can be costly or infeasible.

Finally, we extended our interpretability method to age prediction, an inherently harder and less well-known task. Even in this case, our method uncovered distinct clustering behaviors associated with different age groups and ECG segments. The underrepresented younger patients’ signals tend to be represented with fewer, yet more certain clusters, while older patients display more dispersed and uncertain representations. This suggests that the encoder adapts its internal features in a manner aligned with the amount of data as well as the physiological variation associated with aging, offering a novel perspective on how deep networks encode demographic information from biosignals.

\section{Related work}

Several lines of work have explored explainability in deep learning, both from a general perspective and in the specific context of ECG analysis.

\noindent
\textbf{General explanability methods.} Early methods of feature activations inspection for CNNs~\cite{DeepDream,erhan2009visualizing} provided visual confirmation that deeper layers capture high-level concepts such as objects. Recent developments make it possible to extract semantical information and locate objects from the attention or other internal representations \cite{DINO,LOST,wang2022tokencut}. The disentanglement of the information is performed using PCA \cite{DINO}, SVD \cite{chormai2024disentangled} or neural adaptations of Gaussian mixture models \cite{jacob}. Often, these approaches are guided by saliency maps \cite{LRP, selvaraju_gradcam_2020} thus involving the classifier. Our method focuses solely on disentangling the information extracted by the encoder using a plain $k$-means clustering. The unsupervised nature of our method and the exclusion of the classifiers allow a certain distance between the explanations and the task/predictions, which is necessary for a critical analyses of the training and to reduce task-related biases. One drawback is the necessity to add expert knowledge, which may induce with some biases.

\noindent
\textbf{AI-ECG explanbility.} Explainability methods are commonly employed in AI-ECG studies using deep learning, with heatmaps being one of the most widely used tools to highlight the most relevant regions of the ECG. For example, Grad-CAM has been applied to interpret triage decisions~\cite{vandeleur_automatic_2020}, Chagas disease screening~\cite{jidling_screening_2023}, and the detection of left ventricular systolic dysfunction~\cite{sangha_detection_2023}. Other popular methods include SHapley Additive exPlanations (SHAP)\cite{anand_explainable_2022}; see~\cite{maurer_explainable_2024} for a comprehensive tutorial on explainable AI in biomedical signal processing. Beyond post-hoc analysis, some studies aim for inherently interpretable architectures—for instance, by separating the processing of individual leads~\cite{lu_decoding_2024}. Variational autoencoders have also been used to uncover latent factors underlying ECG morphology~\cite{van_de_leur_improving_2022, kapsecker_disentangled_2025}, offering structured and potentially more interpretable representations.

\section{Conclusion} 

We present a novel interpretability method for CNN-based ECG analysis that, like Grad-CAM~\cite{vandeleur_automatic_2020}, can be applied post hoc to existing models, but offers more expressive insights through intuitive visualizations and quantitative assessments of representation uncertainty. Applied to both classification and regression tasks, it consistently reveals physiologically meaningful patterns, aligning with clinical features and uncovering label-specific structures and model behaviors. The method also highlights uncertainty and data issues, supports clinical validation, and remains robust across hyperparameter settings, making it a practical tool for transparent and trustworthy AI-ECG diagnostics.
port and biomedical discovery.

\section*{Acknowledgments}

FSN and JB supported by British Heart Foundation (BHF) programme grant funding (RG/F/22/110078  and FS/CRTF/24/24624) and Centre of Research Excellence funding (RE/18/4/34215 and RE/24/130023). FSN, KP and JB are supported by the National Institute for Health Research Imperial Biomedical Research Centre.
AHR is partially supported by the eSSENCE strategic collaborative research programme (project ``Digital Biomarkers from the Electrocardiogram using Artificial Intelligence").


\begin{thebibliography}{10}
\bibitem{gbd2016causesofdeathcollaborators_global_2017}
{GBD 2016 Causes of Death Collaborators}, ``Global, regional, and national age-sex specific mortality for 264 causes of death, 1980-2016: a systematic analysis for the {Global} {Burden} of {Disease} {Study} 2016,'' {\em Lancet }, vol.~390, pp.~1151--1210, 2017.

\bibitem{ribeiro_automatic_2020a}
A.~H. Ribeiro, M.~H. Ribeiro, G.~M.~M. Paixão, D.~M. Oliveira, P.~R. Gomes, J.~A. Canazart, M.~P.~S. Ferreira, C.~R. Andersson, P.~W. Macfarlane, W.~Meira~Jr., T.~B. Schön, and A.~L.~P. Ribeiro, ``Automatic diagnosis of the 12-lead {ECG} using a deep neural network,'' {\em Nature Communications}, vol.~11, no.~1, p.~1760, 2020.

\bibitem{jidling_screening_2023}
C.~Jidling, D.~Gedon, T.~B. Schön, C.~D.~L. Oliveira, C.~S. Cardos, A.~M. Ferreira, L.~Giatti, S.~M. Barreto, E.~C. Sabino, A.~L.~P. Ribeiro, and A.~H. Ribeiro, ``Screening for {Chagas} disease from the electrocardiogram using a deep neural network,'' {\em Plos Neglected Tropical Diseases}, vol.~17, no.~7, 2023.

\bibitem{sangha_detection_2023}
V.~Sangha, A.~A. Nargesi, L.~S. Dhingra, B.~J. Mortazavi, A.~H. Ribeiro, C.~A. Brandt, E.~J. Miller, A.~L.~P. Ribeiro, E.~J. Velazquez, H.~M. Krumholz, and R.~Khera, ``Detection of {Left} {Ventricular} {Systolic} {Dysfunction} from {Electrocardiographic} {Images},'' {\em Circulation}, 2023.

\bibitem{von_bachmann_evaluating_2024}
P.~Von~Bachmann, D.~Gedon, F.~K. Gustafsson, A.~H. Ribeiro, E.~Lampa, S.~Gustafsson, J.~Sundström, and T.~B. Schön, ``Evaluating regression and probabilistic methods for {ECG}-based electrolyte prediction,'' {\em Scientific Reports}, vol.~14, no.~15273, 2024.

\bibitem{habineza_end--end_2023}
T.~Habineza, A.~H. Ribeiro, D.~Gedon, J.~A. Behar, A.~L.~P. Ribeiro, and T.~B. Schön, ``End-to-end {Risk} {Prediction} of {Atrial} {Fibrillation} from the 12-{Lead} {ECG} by {Deep} {Neural} {Networks},'' {\em Journal of Electrocardiology}, 2023.

\bibitem{sau_artificial_2024-1}
A.~Sau, L.~Pastika, E.~Sieliwonczyk, K.~Patlatzoglou, A.~H. Ribeiro, K.~A. McGurk, B.~Zeidaabadi, H.~Zhang, K.~Macierzanka, D.~Mandic, E.~Sabino, L.~Giatti, S.~M. Barreto, L.~d.~V. Camelo, I.~Tzoulaki, D.~P. O’Regan, N.~S. Peters, J.~S. Ware, A.~L.~P. Ribeiro, D.~B. Kramer, J.~W. Waks, and F.~S. Ng, ``Artificial intelligence–enabled electrocardiogram for mortality and cardiovascular risk estimation: {An} actionable, explainable and biologically plausible platform,'' 

\bibitem{pastika_artificial_2024}
L.~Pastika, A.~Sau, K.~Patlatzoglou, E.~Sieliwonczyk, A.~H. Ribeiro, K.~A. McGurk, W.~R. Scott, J.~S. Ware, A.~L.~P. Ribeiro, D.~B. Kramer, J.~W. Waks, and F.~S. Ng, ``Artificial intelligence–enabled electrocardiogram for mortality and cardiovascular risk estimation: {An} actionable, explainable and biologically plausible platform,'' {\em npj Digital Medicine}, vol.~7, no.~167, 2024.

\bibitem{koski1996modelling}
A.~Koski, ``Modelling ecg signals with hidden markov models,'' {\em Artificial intelligence in medicine}, vol.~8, no.~5, pp.~453--471, 1996.

\bibitem{andreao2006ecg}
R.~V. Andreao, B.~Dorizzi, and J.~Boudy, ``Ecg signal analysis through hidden markov models,'' {\em IEEE Transactions on Biomedical engineering}, vol.~53, no.~8, pp.~1541--1549, 2006.

\bibitem{boubekki_explaining_2025}
A.~Boubekki, S.~G. Fadel, and S.~Mair, ``Explaining the {Impact} of {Training} on {Vision} {Models} via {Activation} {Clustering},'' Mar. 2025.
\newblock arXiv:2411.19700.

\bibitem{wagner_ptbxl_2020}
P.~Wagner, N.~Strodthoff, R.-D. Bousseljot, D.~Kreiseler, F.~I. Lunze, W.~Samek, and T.~Schaeffter, ``{PTB}-{XL}, a large publicly available electrocardiography dataset,'' {\em Scientific Data}, vol.~7, p.~154, May 2020.

\bibitem{makowski_neurokit2_2021}
D.~Makowski, T.~Pham, Z.~J. Lau, J.~C. Brammer, F.~Lespinasse, H.~Pham, C.~Schölzel, and S.~H.~A. Chen, ``{NeuroKit2}: {A} {Python} toolbox for neurophysiological signal processing,'' {\em Behavior Research Methods}, vol.~53, pp.~1689--1696, 2021.

\bibitem{strodthoff_deep_2020}
N.~Strodthoff, P.~Wagner, T.~Schaeffter, and W.~Samek, ``Deep {Learning} for {ECG} {Analysis}: {Benchmarks} and {Insights} from {PTB}-{XL},'' {\em arXiv:2004.13701 }, 2020.

\bibitem{lima_deep_2021}
E.~M. Lima, A.~H. Ribeiro, G.~M.~M. Paixão, M.~H. Ribeiro, M.~M.~P. Filho, P.~R. Gomes, D.~M. Oliveira, E.~C. Sabino, B.~B. Duncan, L.~Giatti, S.~M. Barreto, W.~Meira, T.~B. Schön, and A.~L.~P. Ribeiro, ``Deep neural network estimated electrocardiographic-age as a mortality predictor,'' {\em Nature Communications}, vol.~12, 2021.

\bibitem{selvaraju_gradcam_2020}
R.~R. Selvaraju, M.~Cogswell, A.~Das, R.~Vedantam, D.~Parikh, and D.~Batra, ``Grad-{CAM}: {Visual} {Explanations} from {Deep} {Networks} via {Gradient}-based {Localization},'' {\em International Journal of Computer Vision}, vol.~128, pp.~336--359, Feb. 2020.

\bibitem{gustafsson_development_2022}
S.~Gustafsson, D.~Gedon, E.~Lampa, A.~H. Ribeiro, M.~J. Holzmann, T.~B. Schön, and J.~Sundström, ``Development and validation of deep learning {ECG}-based prediction of myocardial infarction in emergency department patients,'' {\em Scientific Reports}, vol.~12, p.~19615, 2022.

\bibitem{winkler_association_2019}
J.~K. Winkler, C.~Fink, F.~Toberer, A.~Enk, T.~Deinlein, R.~Hofmann-Wellenhof, L.~Thomas, A.~Lallas, A.~Blum, W.~Stolz, and H.~A. Haenssle, ``Association {Between} {Surgical} {Skin} {Markings} in {Dermoscopic} {Images} and {Diagnostic} {Performance} of a {Deep} {Learning} {Convolutional} {Neural} {Network} for {Melanoma} {Recognition},'' {\em JAMA Dermatology}, 2019.

\bibitem{DeepDream}
K.~Simonyan, A.~Vedaldi, and A.~Zisserman, ``Deep inside convolutional networks: Visualising image classification models and saliency maps,'' {\em arXiv preprint arXiv:1312.6034}, 2013.

\bibitem{erhan2009visualizing}
D.~Erhan, Y.~Bengio, A.~Courville, and P.~Vincent, ``Visualizing higher-layer features of a deep network,'' tech. rep., University of Montreal, 2009.

\bibitem{DINO}
M.~Caron, H.~Touvron, I.~Misra, H.~J{\'e}gou, J.~Mairal, P.~Bojanowski, and A.~Joulin, ``Emerging properties in self-supervised vision transformers,'' in {\em Proceedings of the IEEE/CVF International Conference on Computer Vision}, pp.~9650--9660, 2021.

\bibitem{LOST}
O.~Sim\'eoni, G.~Puy, H.~V. Vo, S.~Roburin, S.~Gidaris, A.~Bursuc, P.~P\'erez, R.~Marlet, and J.~Ponce, ``Localizing objects with self-supervised transformers and no labels,'' in {\em Proceedings of the British Machine Vision Conference}, 2021.

\bibitem{wang2022tokencut}
Y.~Wang, X.~Shen, S.~X. Hu, Y.~Yuan, J.~L. Crowley, and D.~Vaufreydaz, ``Self-supervised transformers for unsupervised object discovery using normalized cut,'' in {\em Conference on Computer Vision and Pattern Recognition}, (New Orleans, LA, USA), 2022.

\bibitem{chormai2024disentangled}
P.~Chormai, J.~Herrmann, K.-R. M{\"u}ller, and G.~Montavon, ``Disentangled explanations of neural network predictions by finding relevant subspaces,'' {\em IEEE Transactions on Pattern Analysis and Machine Intelligence}, 2024.

\bibitem{jacob}
J.~Kauffmann, M.~Esders, L.~Ruff, G.~Montavon, W.~Samek, and K.-R. M{\"u}ller, ``From clustering to cluster explanations via neural networks,'' {\em IEEE Transactions on Neural Networks and Learning Systems}, 2022.

\bibitem{LRP}
S.~Bach, A.~Binder, G.~Montavon, F.~Klauschen, K.-R. M{\"u}ller, and W.~Samek, ``On pixel-wise explanations for non-linear classifier decisions by layer-wise relevance propagation,'' {\em PloS ONE}, vol.~10, no.~7, p.~e0130140, 2015.

\bibitem{vandeleur_automatic_2020}
R.~R. van~de Leur, L.~J. Blom, E.~Gavves, I.~E. Hof, J.~F. van~der Heijden, N.~C. Clappers, P.~A. Doevendans, R.~J. Hassink, and R.~van Es, ``Automatic {Triage} of 12‐{Lead} {ECGs} {Using} {Deep} {Convolutional} {Neural} {Networks},'' {\em Journal of the American Heart Association}, vol.~9, 2020.

\bibitem{anand_explainable_2022}
A.~Anand, T.~Kadian, M.~K. Shetty, and A.~Gupta, ``Explainable {AI} decision model for {ECG} data of cardiac disorders,'' {\em Biomedical Signal Processing and Control}, vol.~75, p.~103584,  2022.

\bibitem{maurer_explainable_2024}
M.~C. Maurer, J.~M. Metsch, P.~Hempel, T.~Bender, N.~Spicher, and A.-C. Hauschild, ``Explainable {Artificial} {Intelligence} on {Biosignals} for {Clinical} {Decision} {Support},'' in {\em Proceedings of the 30th {ACM} {SIGKDD} {Conference} on {Knowledge} {Discovery} and {Data} {Mining}}, (Barcelona Spain), pp.~6597--6604, ACM, 2024.

\bibitem{lu_decoding_2024}
L.~Lu, T.~Zhu, A.~H. Ribeiro, L.~Clifton, E.~Zhao, J.~Zhou, A.~L.~P. Ribeiro, Y.-T. Zhang, and D.~A. Clifton, ``Decoding 2.3 {Million} {ECGs}: {Interpretable} {Deep} {Learning} for {Advancing} {Cardiovascular} {Diagnosis} and {Mortality} {Risk} {Stratification},'' {\em European Heart Journal - Digital Health}, 2024.

\bibitem{van_de_leur_improving_2022}
R.~R. van~de Leur, M.~N. Bos, K.~Taha, A.~Sammani, M.~W. Yeung, S.~van Duijvenboden, P.~D. Lambiase, R.~J. Hassink, P.~van~der Harst, P.~A. Doevendans, D.~K. Gupta, and R.~van Es, ``Improving explainability of deep neural network-based electrocardiogram interpretation using variational auto-encoders,'' {\em European Heart Journal - Digital Health}, vol.~3, pp.~390--404, 2022.

\bibitem{kapsecker_disentangled_2025}
M.~Kapsecker, M.~C. Möller, and S.~M. Jonas, ``Disentangled representational learning for anomaly detection in single-lead electrocardiogram signals using variational autoencoder,'' {\em Computers in Biology and Medicine}, vol.~184, p.~109422, 2025.

\end{thebibliography}

\clearpage
\onecolumn
\appendix

\begin{figure*}[!h]
    {\centering \small
    \begin{tabular*}{\linewidth}{@{\extracolsep{\fill}}cc}
    Explanation's Uncertainty & GCAM Importance \vspace{1em}\\ 
    \includegraphics[width=.5\linewidth]{img/ECG_Examples_uncert.png} & 
    \includegraphics[width=.5\linewidth]{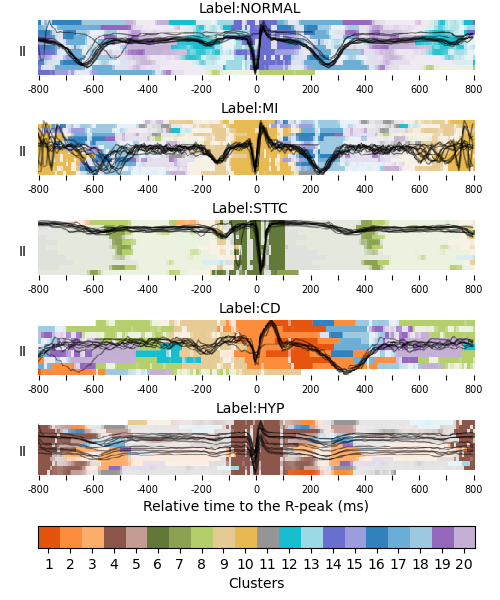}
    \end{tabular*}}
    \caption{Comparison of the explanations' certainty and of the GCAM importance on examples of Figure~\ref{fig:example}.}
    \label{fig-apx:uncert}
\end{figure*}
\begin{figure*}[!h]
    \centering \small
    \begin{tabular*}{\linewidth}{@{\extracolsep{\fill}}cc}
    K=5 & K=10 \\
    \includegraphics[width=.5\linewidth]{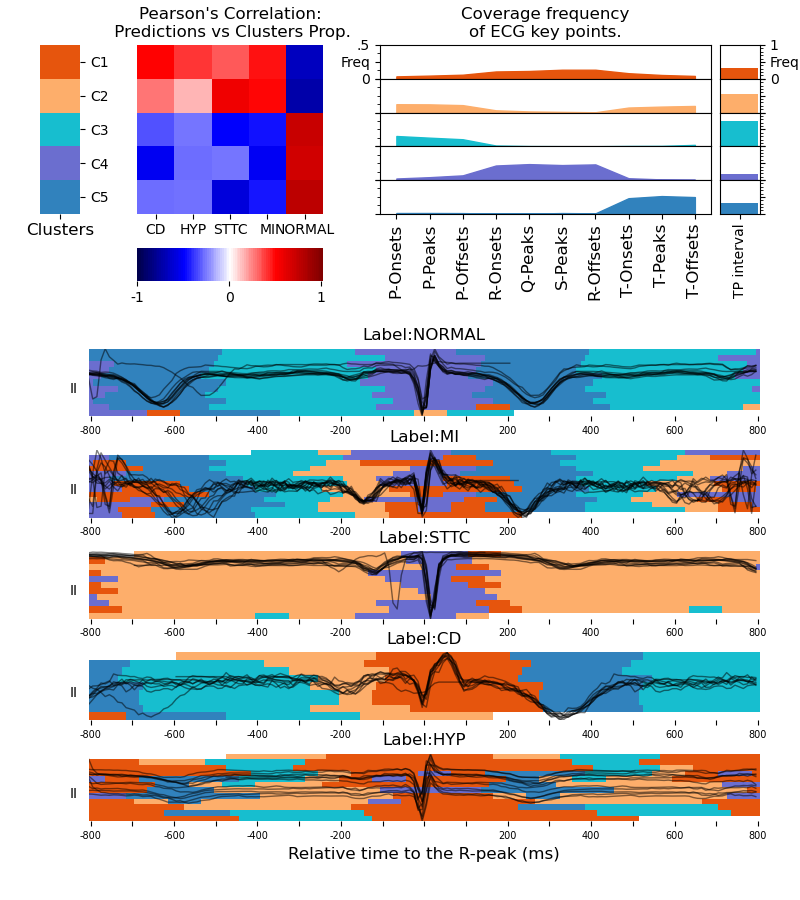} & 
    \includegraphics[width=.5\linewidth]{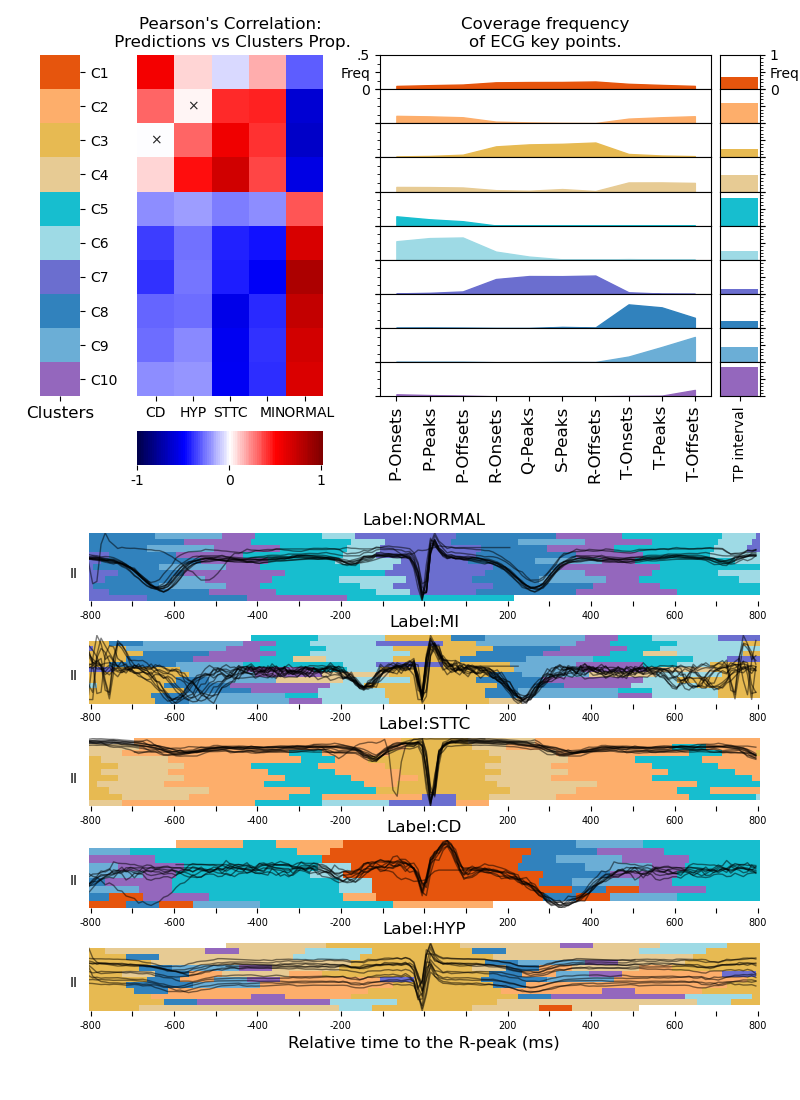}
    \end{tabular*}
    \caption{Explanations' alignment with predicted labels and ECG key points and some examples (Figure~\ref{fig:example}) for different number of clusters.}
    \label{fig-apx:ablk}
\end{figure*}
\begin{figure*}[!h]
    \centering \small
    \begin{tabular*}{\linewidth}{@{\extracolsep{\fill}}cc}
    K=15 & K=20 \\
    \includegraphics[width=.5\linewidth]{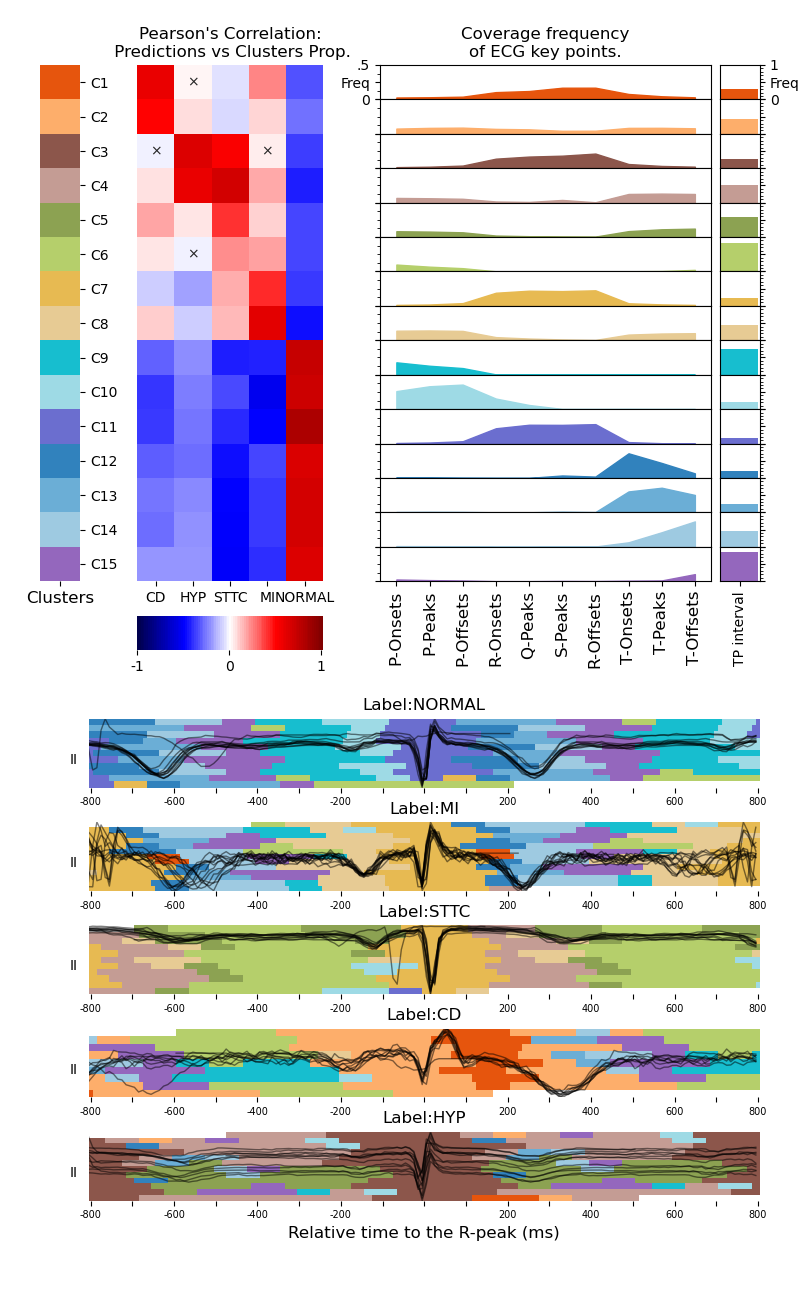} &
    \includegraphics[width=.5\linewidth]{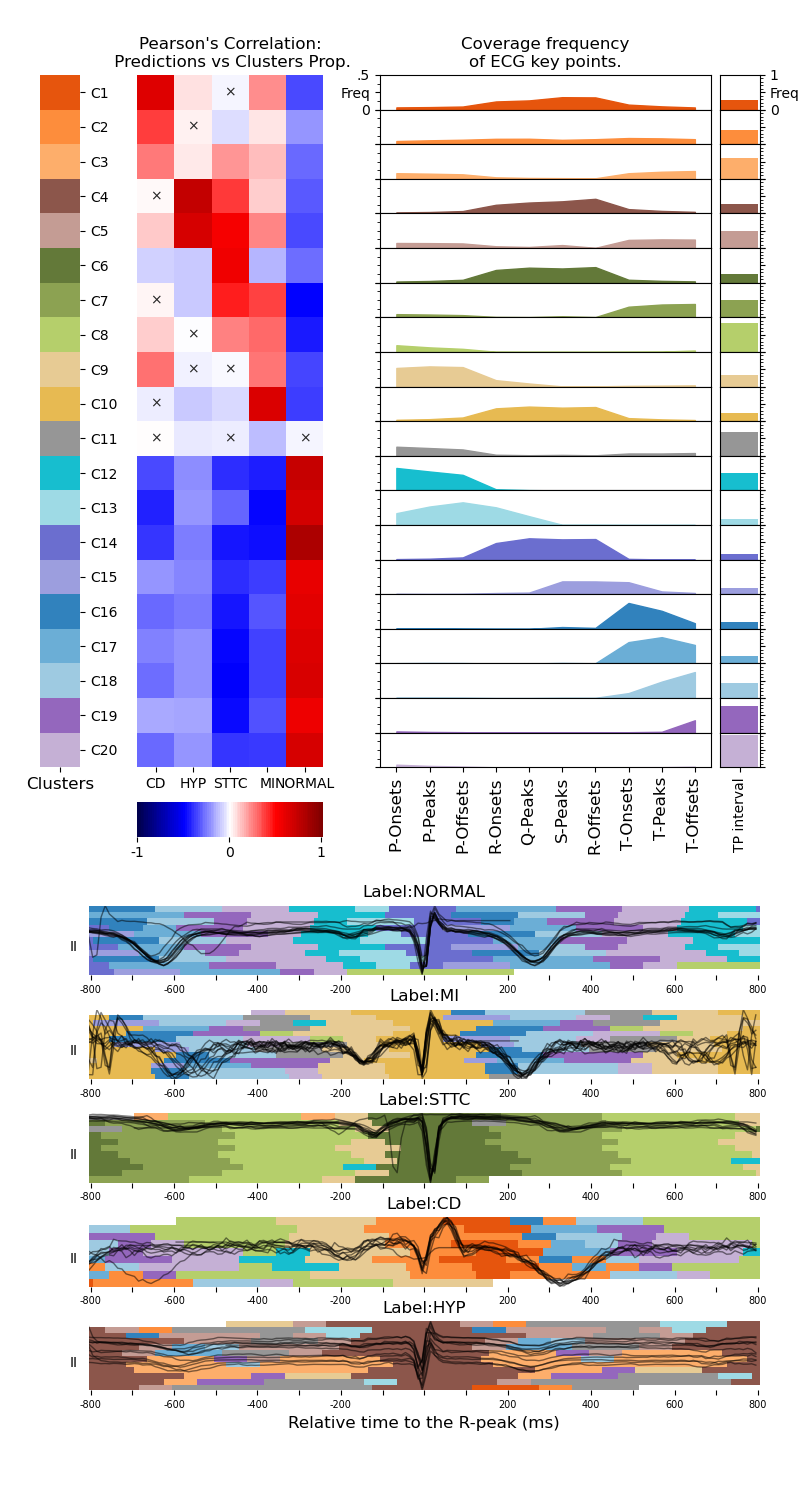}
    \end{tabular*}
    \caption{Explanations' alignment with predicted labels and ECG key points and some examples (Figure~\ref{fig:example}) for different number of clusters.}
    \label{fig-apx:ablk2}
\end{figure*}

\begin{figure*}[!h]
    \centering \small
    a. Expert label error with concordant model predictions \\
    \includegraphics[width=.95\linewidth]{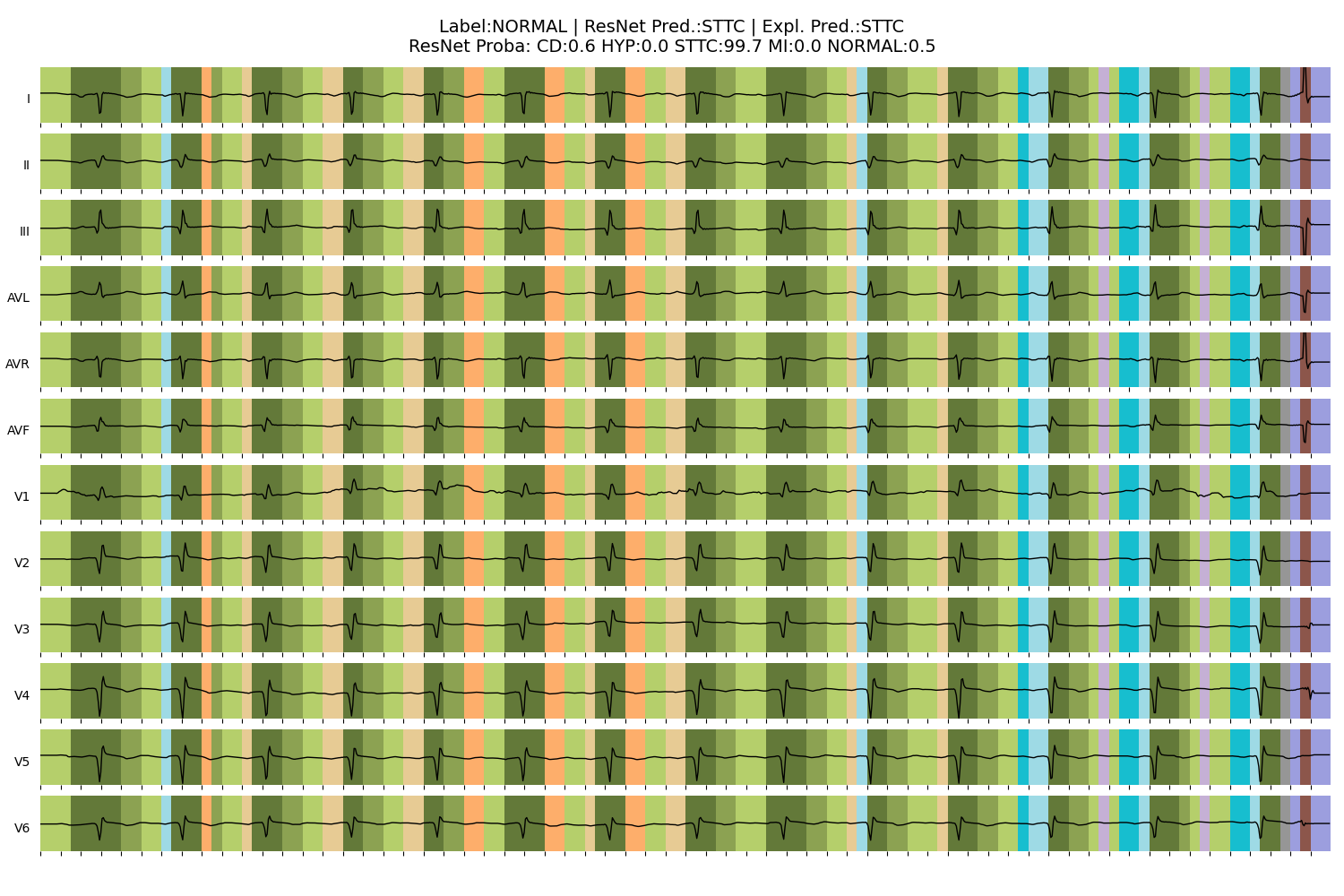}\\ \vspace{.5em}
    \centering \small
    b. Expert label error with discordant model predictions \\
    \includegraphics[width=.95\linewidth]{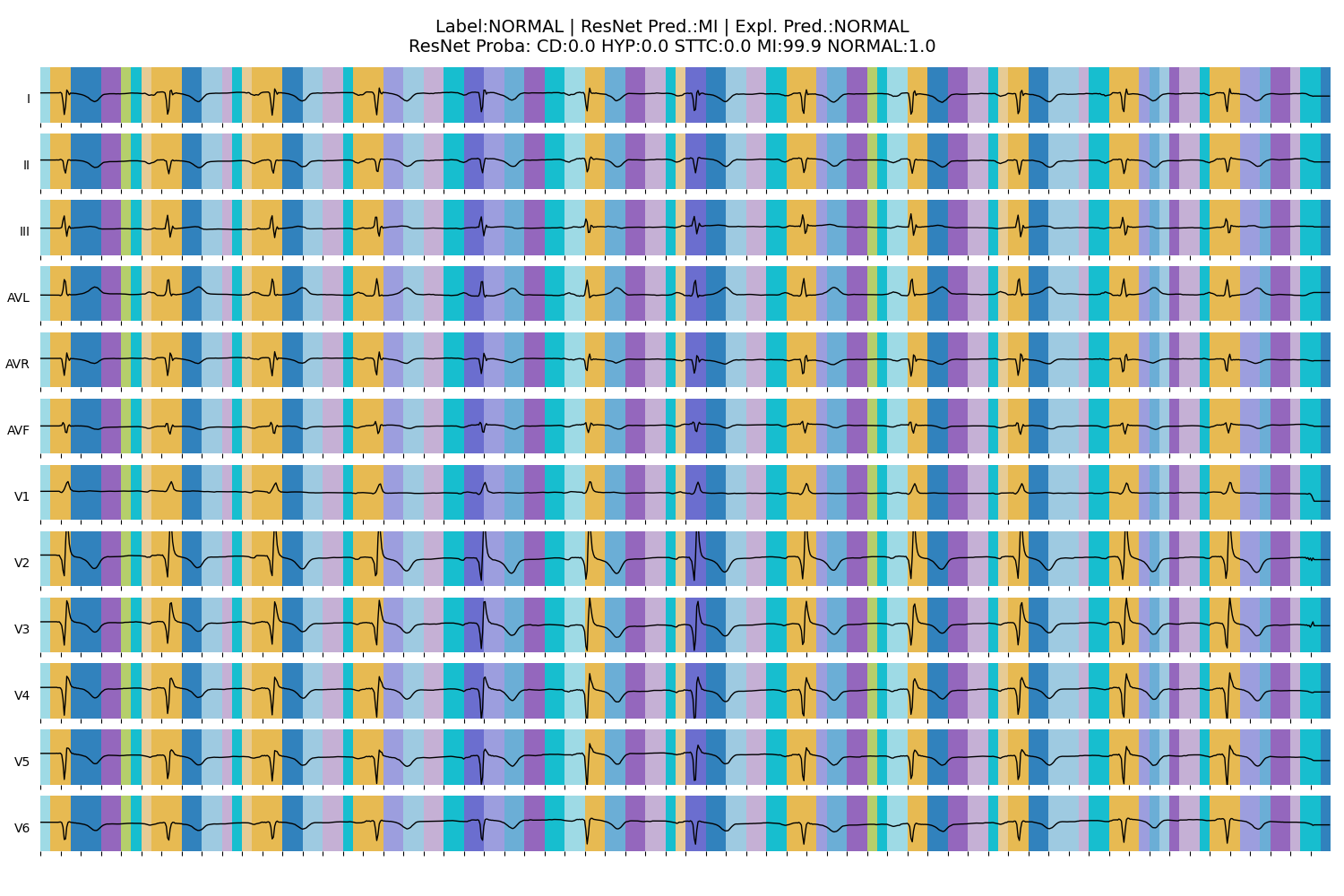}
    \caption{12 Lead ECG misclassifications with explanations of Figure~\ref{fig:misclass}}
    \label{fig:misclass_full_1}
\end{figure*}
\begin{figure*}[!h]
    \centering \small

    c. Limitations of global labels for transient ECG morphologies \\
    \includegraphics[width=.95\linewidth]{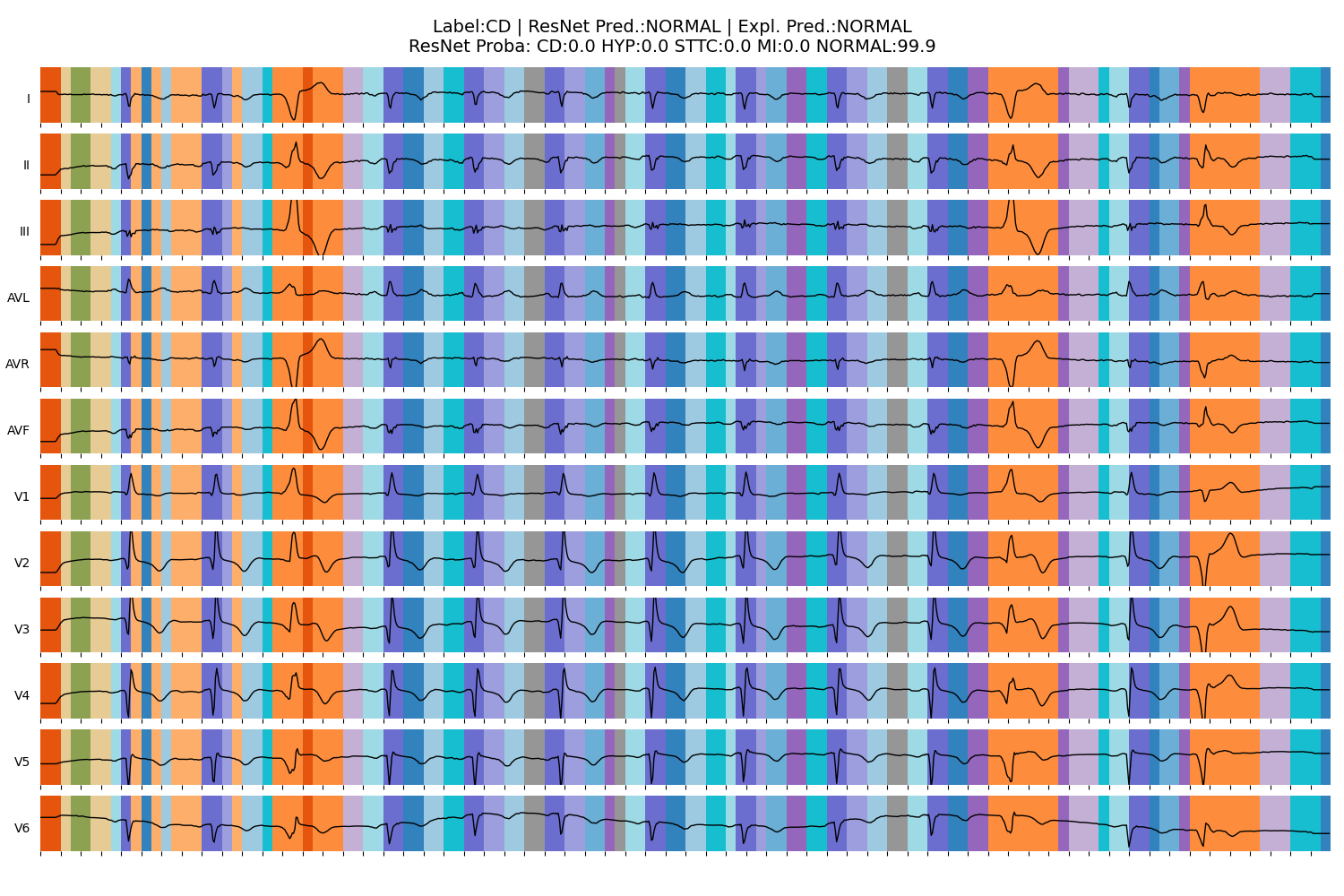}
    \caption{12 Lead ECG misclassifications with explanations of Figure~\ref{fig:misclass}}
    \label{fig:misclass_full_2}
\end{figure*}

\end{document}